\documentclass[letterpaper, 10 pt, conference]{ieeeconf}  

\IEEEoverridecommandlockouts                              

\usepackage{cite}
\usepackage{subfigure}
\usepackage{graphics} 
\usepackage{epsfig} 
\usepackage{times} 
\usepackage{amsmath} 
\usepackage{amssymb}  
\usepackage{booktabs}
\usepackage{multirow}
\usepackage{makecell}
\usepackage{tabularx}
\usepackage{siunitx}

\title{\LARGE \bf
SignNav: Leveraging Signage for Semantic Visual Navigation\\ in Large-Scale Indoor Environments
}

\author{
Jian Sun$^1$, Yuming Huang$^1$, He Li$^1$, Shuqi Xiao$^1$, Shenyan Guo$^1$, \\Maani Ghaffari$^2$, Qingbiao Li$^1$, Chengzhong Xu$^1$, \textit{Fellow}, \textit{IEEE}, Hui Kong$^{1, \dagger}$%
\thanks{$^{\dagger}$ denotes the corresponding author: {\tt\small huikong@um.edu.mo}}%
\thanks{$^{1}$ Faculty of Science and Technology, University of Macau}%
\thanks{$^{2}$ Department of Robotics, University of Michigan}%
}

\begin{document}
\bstctlcite{IEEEexample:BSTcontrol}
\maketitle
\thispagestyle{empty}
\pagestyle{empty}
\begin{abstract}
Humans routinely leverage semantic hints provided by signage to navigate to destinations within novel Large-Scale Indoor (LSI) environments, such as hospitals and airport terminals. However, this capability remains underexplored within the field of embodied navigation. This paper introduces a novel embodied navigation task, \textbf{SignNav}, which requires the agent to interpret semantic hint from signage and reason about the subsequent action based on current observation. To facilitate research in this domain, we construct the \textbf{LSI-Dataset} for the training and evaluation of various SignNav agents. Dynamically changing semantic hints and sparse placement of signage in LSI environments present significant challenges to the SignNav task. To address these challenges, we propose the Spatial-Temporal Aware Transformer (START) model for end-to-end decision-making. The spatial-aware module grounds the semantic hint of signage into physical world, while the temporal-aware module captures long-range dependencies between historical states and current observation. Leveraging a two-stage training strategy with Dataset Aggregation (DAgger), our approach achieves state-of-the-art performance, recording an 80\% Success Rate (SR) and 0.74 NDTW on val-unseen split. Real-world deployment further demonstrates the practicality of our method in physical environment without pre-built map.
\end{abstract}
\section{INTRODUCTION}
Navigating Large-Scale Indoor (LSI) environments, such as airport terminals and hospitals, remains a fundamental challenge for embodied agents. While humans intuitively follow high-level semantic cues, specifically signage, to traverse novel spaces, replicating this capability in robots is an elusive goal. Operating in such settings requires not only perceiving environmental semantics but also reasoning over long-horizon actions based on dynamic visual cues. Although numerous embodied navigation tasks have been proposed in recent years \cite{anderson2018evaluation}, signage-driven navigation within LSI environments remains significantly underexplored.
\begin{figure}[tp]
    \centering
    \includegraphics[width=0.85\linewidth]{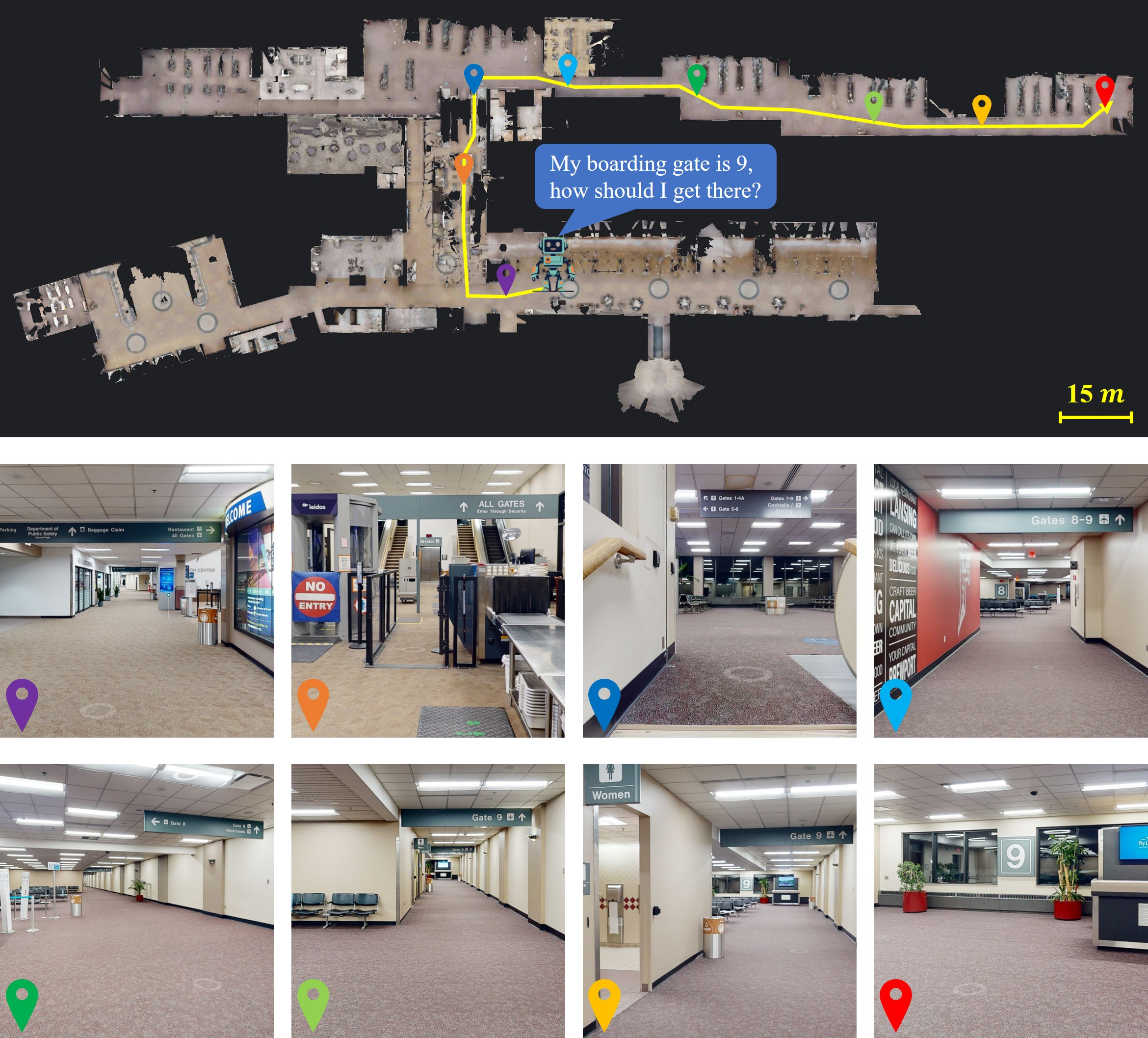}
    \caption{\textbf{Capital Region International Airport in Clinton County, Michigan}. When humans first arrive at an airport terminal, they routinely leverage the semantic hints of signage to find their boarding gates. We formulate this problem as an embodied navigation task \textbf{SignNav} and build the \textbf{LSI-Dataset} to promote the solution of SignNav.
    }
    \label{SignNav_4}
    \vspace{-8pt}
\end{figure}

Driven by photo-realistic datasets and simulators \cite{chang2017matterport3d, savva2019habitat}, paradigms like Object Navigation (ObjectNav) \cite{batra2020objectnav} and Vision-and-Language Navigation (VLN) \cite{anderson2018vision} have advanced significantly. However, ObjectNav is typically constrained to a finite set of target categories and lacks directional reasoning, while VLN relies on granular, step-by-step instructions that are rarely available in public spaces. Recent zero-shot approaches leveraging Large Language Models (LLMs) and Vision-Language Models (VLMs) \cite{majumdar2022zson, yokoyama2024vlfm, chen20232, zhou2024navgpt} improve robustness but are primarily validated in small-scale household settings, falling short in the expansive and semantically complex context of LSI environments.

In public spaces, signage serves as an implicit, omnipresent instruction set for human navigation, providing directional semantics that bridge current locations and destinations (Fig. \ref{SignNav_4}). To formalize this paradigm for embodied agents, we introduce \textbf{SignNav}, a novel navigation task where robots must interpret visual signage to deduce sequential actions based on current observations and historical context. To facilitate research in this domain, we construct the \textbf{LSI-Dataset}, comprising 20 signage-labeled LSI environments and a large-scale collection of automatically generated trajectory-action episodes. While intuitive for humans, SignNav poses unique challenges for robots. (1) \textbf{Dynamic Semantic Grounding:} Unlike ObjectNav or VLN, where navigation guidance is fixed initially, signage semantics and spatial configurations evolve continuously. Agents must dynamically ground these transient hints into egocentric observations to inform decision-making. (2) \textbf{Sparse Observability:} Signage is sparsely distributed in LSI environments. The agent must maintain a temporal memory of previously observed hints to reason effectively when explicit guidance is temporarily invisible.

To address these challenges, we propose the \textbf{S}patial-\textbf{T}emporal \textbf{A}wa\textbf{R}e \textbf{T}ransformer (\textbf{START}), an end-to-end decision-making model (Fig. \ref{SignNav_3}). START features a spatial-aware module that grounds semantic hints into current visual observations, and a temporal-aware module that leverages historical states to capture long-range dependencies. Trained via a two-stage strategy (teacher forcing followed by DAgger \cite{ross2011reduction}), START achieves state-of-the-art performance on the SignNav task, yielding an 80\% SR and 0.74 NDTW on the val-unseen split, with an average trajectory error of merely 0.26 m over long-horizon navigations. Finally, we demonstrate the practical efficacy of START through mapless, real-world deployments on a DeepRobotics Lite2 robot in an office building.
\section{Related Works}
\subsection{Object Goal Navigation}
Object Goal Navigation (ObjectNav) tasks an agent with locating an instance of a specific object category in an unseen environment. This paradigm relies on the premise that leveraging semantic priors enables efficient search and localization \cite{batra2020objectnav}. Learning-based approaches to train agents with semantic navigation abilities have typically leveraged reinforcement learning \cite{deitke202, yadav2023ovrl}, learning from demonstration \cite{ramrakhya2023pirlnav}, or prediction of semantic top-down maps \cite{chaplot2020object, ramakrishnan2022poni} on which waypoint planners can be used. However, standard ObjectNav is fundamentally limited by the closed-set of object categories present in training data. While recent zero-shot approaches \cite{majumdar2022zson, gadre2023cows}, particularly those leveraging LLMs and VLMs \cite{dorbala2023can, yu2023l3mvn, yokoyama2024vlfm}, have expanded this scope to open-set targets, they are predominantly validated in small-scale household settings. Crucially, ObjectNav assumes the target is visually recognizable from a distance or upon discovery. In contrast, our work addresses navigation in LSI environments, such as office buildings and hospitals where the destination is often not directly visible, necessitating the interpretation of intermediate semantic hints provided by signage.

\subsection{Vision-and-Language Navigation}
Vision-and-Language Navigation (VLN) is a multi-modal task where agents follow natural language instructions (e.g., \textit{``Turn left, then stop by the sofa"}) to navigate through real or simulated environments.
One of the most representative works in VLN is VLN-CE benchmark \cite{krantz_vlnce_2020}. Methods in this domain range from predicting control commands directly \cite{chen2021topological, georgakis2022cross} to hierarchical approaches selecting subgoals \cite{hong2022bridging, krantz2021waypoint}. Subsequent advancements leveraged vision-language pre-trained models \cite{su2019vl, tan2019lxmert} or task-specific pre-training frameworks \cite{hao2020towards, majumdar2020improving} to enhance performance. More recently, the advancements in LLMs have shifted VLN from task-specific optimization to open-domain instruction generalization \cite{zheng2024towards, long2024instructnav, zhou2024navgpt}. Despite these successes, VLN inherently depends on granular, step-by-step guidance provided by a human or an oracle, which might be unavailable in real-world deployment. Our work diverges from this paradigm by treating signage as an \textit{embedded, dynamic instruction set}. This allows the agent to navigate instantly by extracting explicit semantic hints directly from the environment.

\subsection{Signage-Aware Navigation}
While signage is crucial for wayfinding in human-centric environments by abstracting spatial information into symbolic representations \cite{gibson2009wayfinding}, its integration into embodied navigation remains underexplored. Prior research primarily addresses isolated perception tasks, such as sign detection \cite{almeida2019indoor, wang2013detecting}, semantic parsing \cite{liang2020perceiving}, or global localization \cite{chen2025signage, zimmerman2025signloc}. Recently, \textit{Sign Language} \cite{agrawal2025sign} proposed a VLM-based baseline for signage understanding, but it lacks closed-loop navigation control. The most closely related work, \textit{No Map No Problem} \cite{liang2020no}, employs a rule-based framework for mapless navigation by parsing text and directions. Similarly, concurrent works \cite{cao2026follow, lee2026iros} explore signage-based navigation but rely on modular pipelines that pair LLMs with traditional planners. In contrast to these rule-based or modular methods, we formulate SignNav as an end-to-end learning problem. We propose a \textit{Spatial-Temporal Aware Transformer} that directly maps visual observations and semantic hints to physical actions via spatial grounding and temporal reasoning. To our knowledge, we are the first to formally define the end-to-end SignNav task and provide a dedicated dataset to support it.
\section{Leveraging Signage for Semantic Visual Navigation}
This work bridges the gap between high-level signage semantics and low-level navigation actions. By decoupling decision making from raw signage parsing, we assume access to structured semantic outputs, specifically, the mapping between navigational goals and directional indicators. Consequently, we abstract visual signage into directional arrows for navigation. Given that recent zero-shot VLM method \cite{agrawal2025sign} demonstrate robust capabilities in signage perception, this abstraction allows us to focus exclusively on downstream policy learning.

\subsection{Task Definition}
We formulate the \textbf{SignNav} task as a Partially Observable Markov Decision Process (POMDP). The objective is for an embodied agent to navigate continuously in 3D environment to reach a semantic target location (e.g., \textit{``Boarding Gate 9"}) by interpreting visual semantic hints extracted from signage. Unlike ObjectNav or VLN, the agent must rely on directional hints (e.g., arrows pointing to the target area) dynamically detected in the environment.

\noindent\textbf{Action Space.} The agent interacts with the environment through a discrete action space $a_t \in \mathcal{A}$ consisting of four low-level navigable commands: \textsf{Move Forward} ($0.25\text{m}$), \textsf{Turn Left} ($15^\circ$), \textsf{Turn Right} ($15^\circ$), and \textsf{Stop}. This setup aligns with the environment standards established in VLN-CE \cite{krantz_vlnce_2020}.

\noindent\textbf{Observation Space.} At each time step $t$, the agent receives a composite observation tuple $(o_t, h_t, b_t)$. Here, $o_t = \{o^r_t, o^d_t\}$ represents the egocentric perception, comprising an RGB image $o^r_t$ and a depth image $o^d_t$. Crucially, to simulate the attention mechanism of reading signs, we introduce the visual semantic hint $h_t$, defined as the local image region (crop) enclosed by the bounding box $b_t$ of the detected directional arrow relevant to the target.

\noindent\textbf{Policy Learning.} Due to the partial observability of the environment, the agent must maintain a history of past states to resolve ambiguities (e.g., remembering a sign seen before turning at a corner). We define the history at step $t$ as $\mathcal{H}_t \triangleq \{(o_0, h_0, a_0), \ldots, (o_{t-1}, h_{t-1}, a_{t-1})\}$. The goal is to learn a policy $\pi$, parameterized by $\Theta$, that maps the current observation and history to a probability distribution over the action space:
\begin{equation}
    a_t \sim \pi(a_t \mid \mathcal{H}_t, o_t, h_t, b_t;\Theta)
\end{equation}
The agent executes the sampled action $a_t$, transitioning to a new state $s_{t+1}$ and receiving a new observation, continuing until the \textsf{Stop} action is invoked or a maximum step limit is reached.
\begin{figure}[tp]
    \centering
    \includegraphics[width=0.85\linewidth]{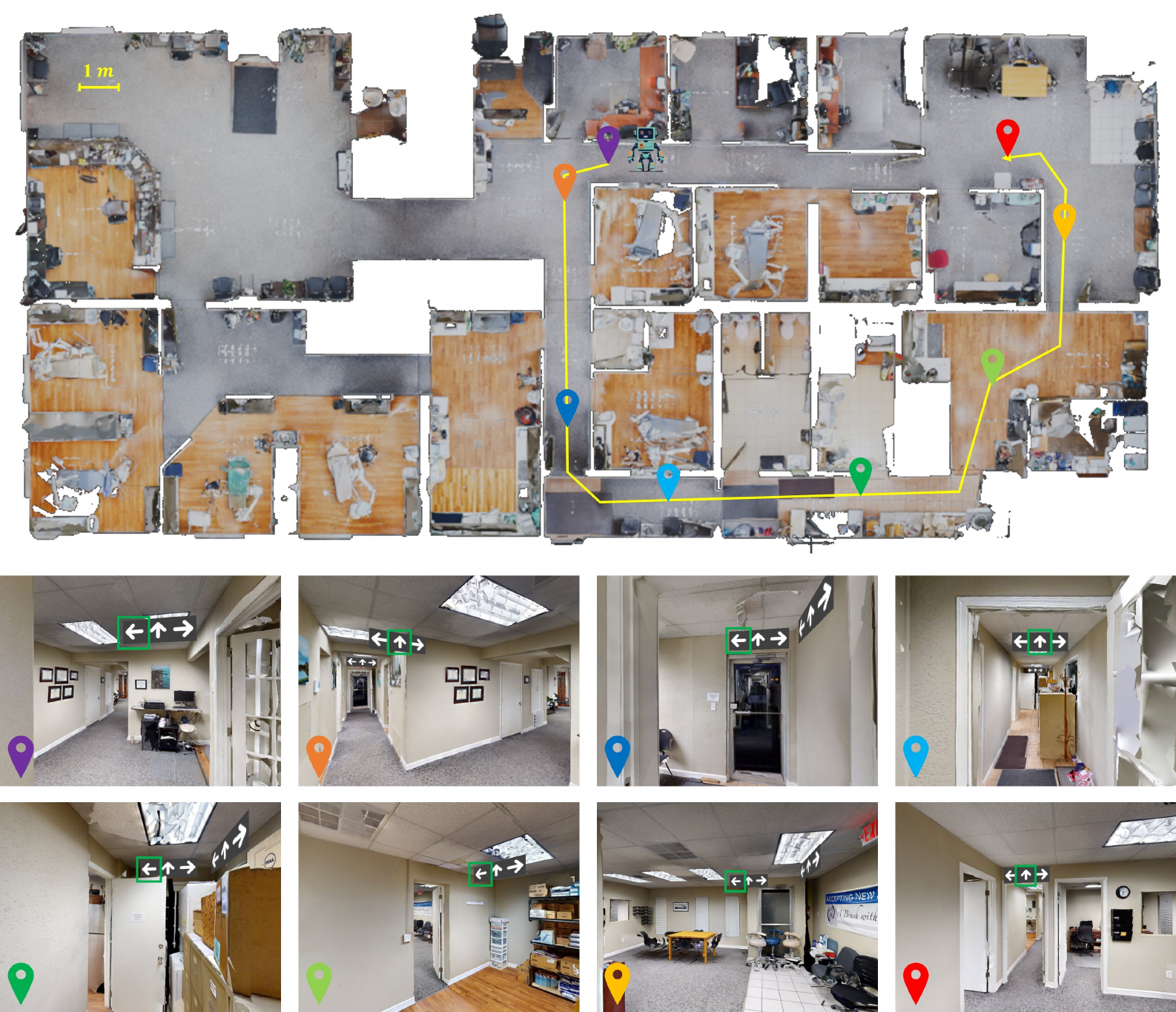}
    \caption{\textbf{An episode example in a hospital scene.} The agent is required to make action decisions according to the semantic hints of signage (detected directional arrows) and finally arrive at the target location.}
    \label{SignNav_6}
    \vspace{-8pt}
\end{figure}

\subsection{LSI-Dataset Creation.}
Existing embodied navigation datasets \cite{chang2017matterport3d, ramakrishnan2021habitat} predominantly feature small-scale household environments, which fail to capture the spatial complexity and scale of public infrastructure. To bridge this gap, we introduce the \textbf{L}arge-\textbf{S}cale \textbf{I}ndoor (LSI) Dataset. We collect 20 high-fidelity 3D scans from open-source repositories \cite{chang2017matterport3d, xia2018gibson, armeni20163d}, specifically selecting public spaces such as hospitals and office complexes. The LSI-Dataset features environments with an average floor area of $1000\,\text{m}^2$, approximately three times larger than standard household datasets. These scenes present unique challenges for SignNav, including long corridors, open halls, and complex opological structures that require sustained directional guidance.

\noindent\textbf{Signage Injection.}
To scale data collection without manual annotation, we developed an automated pipeline using Blender to procedurally inject signage into the 3D meshes. We identify decision points (e.g., intersections, corners) within the scene geometry and instantiate signage assets aligned with potential navigation paths. Fig. \ref{SignNav_6} illustrates the rendered effect in a hospital scene. \textit{Design Choice}: In this work, we deliberately utilize signage containing only directional arrows, excluding textual information. This design choice decouples the challenge of \textit{spatial reasoning} (mapping a visual arrow to a physical action) from \textit{semantic matching} (OCR and text association). By isolating the directional component, we focus on the fundamental problem of how an agent perceives and grounds geometric cues into continuous control.

\noindent\textbf{Episode Generation.}
Unlike VLN, where instructions are static and global, SignNav relies on dynamic, local visual cues that evolve as the agent moves. To generate training and evaluating episodes, we first construct a navigable connectivity graph $\mathcal{G} = (\mathcal{V}, \mathcal{E})$ for each scene using Poisson Disk Sampling \cite{PoissonDiskSampling}, which ensures the waypoints are centered within navigable space and maximize clearance from obstacles. We sample start and goal pairs from $\mathcal{V}$ and plan the shortest path using Dijkstra's algorithm. Crucially, to simulate realistic human-like movement, we apply spline-based smoothing to the discrete path. To automatically annotate the semantic hints (i.e., directional arrows) for the signage, we determine the appropriate arrow direction by leveraging the local curvature of the smoothed trajectory and the agent's heading relative to the injected signage.

Formally, an episode is recorded by an oracle agent executing the planned path. At each time step $t$, we record the RGB-D observation $o_t$, the bounding-box crop of the labeled directional hint $h_t$, and the oracle action $a^\ast_t$. The resulting episode is formulated as:
\begin{equation}
    \tau = \left\{ (o_t, h_t, b_t, a^\ast_t) \right\}_{t=0}^{T}
\end{equation}
where $T$ denotes the horizon of the episode. This automatic data generation pipeline allows us to collect a massive corpus of diverse trajectories, enabling the agent to learn the mapping from visual hints to navigable actions effectively.
\begin{figure*}[tp]
    \centering
    \includegraphics[width=0.8\linewidth]{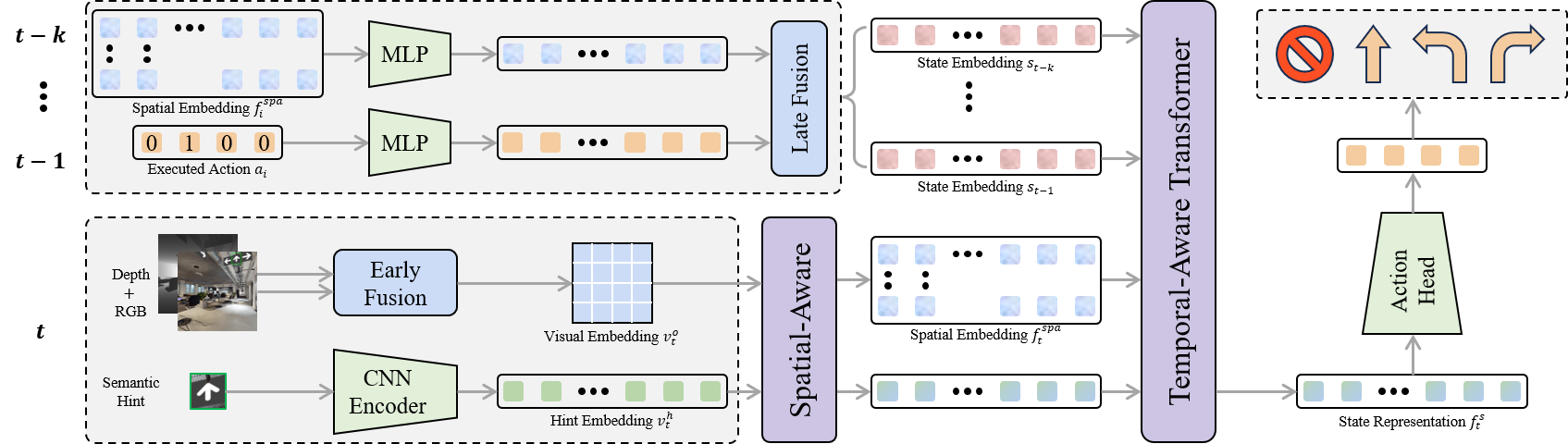}
    \caption{\textbf{The architecture of our Spatial-Temporal Aware Transformer (START) model for the SignNav task.} START uses a spatial-aware module for grounding the semantic hint of sigange into egocentric observation, and a temporal-aware module for capturing long-range dependencies between the historical states and current observation.}
    \label{SignNav_3}
    \vspace{-8pt}
\end{figure*}

\section{Spatial-Temporal Aware Transformer}
Figure \ref{SignNav_3} illustrates the hierarchical architecture of our \textbf{START} model, which interpret visual semantic hints through two distinct stages: spatial grounding of current observations and temporal reasoning over historical states. At current time step $t$, the agent receives heterogeneous inputs: the egocentric RGB-D observation $o_t$ and the cropped semantic hint $h_t$. First, the RGB and depth modalities are merged via early fusion to produce the visual embedding $v_t^o$, while the hint is encoded into $v_t^h$ via a CNN encoder. To resolve the grounding relationship between the directional hint and the environment geometry, these features are processed by a \textit{Spatial-Aware Transformer}, yielding the spatial embedding $f_t^{spa}$. Simultaneously, to maintain trajectory consistency, we construct a history memory bank. Historical spatial embeddings and executed actions from previous steps ($t-k$ to $t-1$) are integrated via late fusion to form a sequence of state embeddings $\{s_{t-k}, \dots, s_{t-1}\}$. This historical sequence, combined with the current spatial and hint features, is fed into the \textit{Temporal-Aware Transformer}. This module captures long-range dependencies and updates the current state representation $s_t$. Finally, an \textit{Action Head} projects $s_t$ to predict the probability distribution $p_t^a$ over the next navigable action.

\subsection{Input Embeddings}
\textbf{Observation Embedding.} Given the RGB-D observation $o_t = \{o^r_t,o^d_t\}$ at time step $t$, we first separately encode the RGB image $o^r_t$ and depth image $o^d_t$ using two lightweight CNN encoders to explicitly extract low-resolution visual features. These two CNN encoders share the same model architecture but with different input channels. The encoded RGB and depth visual features, $\{v^r_t, v^d_t\}$, are concatenated along the channel dimension followed by layer normalization to form the intermediate visual embedding:
\begin{equation}
    v^o_t = {\rm LayerNorm}(v^r_t + v^d_t)
\end{equation}
This fusion strategy effectively integrates geometric structure with semantic texture. The resulting $v^o_t$ serves as the grid-based input for the subsequent \textit{Spatial-Aware Transformer}.

\textbf{Hint Embedding.} To preserve fine-grained visual cues, we bypass textual descriptions (e.g., ``turn right'') and directly leverage the visual feature of the navigation hint. Specifically, given the cropped hint image $h_t$ and its bounding box coordinates $b_t = [x_{min}, y_{min}, x_{max}, y_{max}]$, we construct a semantic-spatial hint embedding $v^h_t$. The visual feature $f^h_t \in \mathbb{R}^{2048}$ is extracted via a ResNet-50 \cite{he2016deep} pre-trained on ImageNet. For the coordinates feature $f^b_t$, we apply $f^b_t = [\frac{x_{min}}{W}, \frac{y_{min}}{H}, \frac{x_{max}}{W}, \frac{y_{max}}{H}, \frac{w\cdot h}{W\cdot H}]$. The final hint embedding $v^h_t$ is obtained by projecting and fusing these two features:
\begin{equation}
    v^h_t = {\rm LayerNorm}(f^h_t \boldsymbol{W}^h) + {\rm LayerNorm}(f^b_t \boldsymbol{W}^p)
\end{equation}
where $\boldsymbol{W}^h$ and $\boldsymbol{W}^p$ are learnable projection matrices that map the visual and coordinates features into a common latent space.

\subsection{Spatial-Aware Transformer}
\begin{figure}[tp]
    \centering
    \subfigure[]
    {
        \begin{minipage}[b]{0.26\linewidth}
            \centering
            \includegraphics[width=1\linewidth]{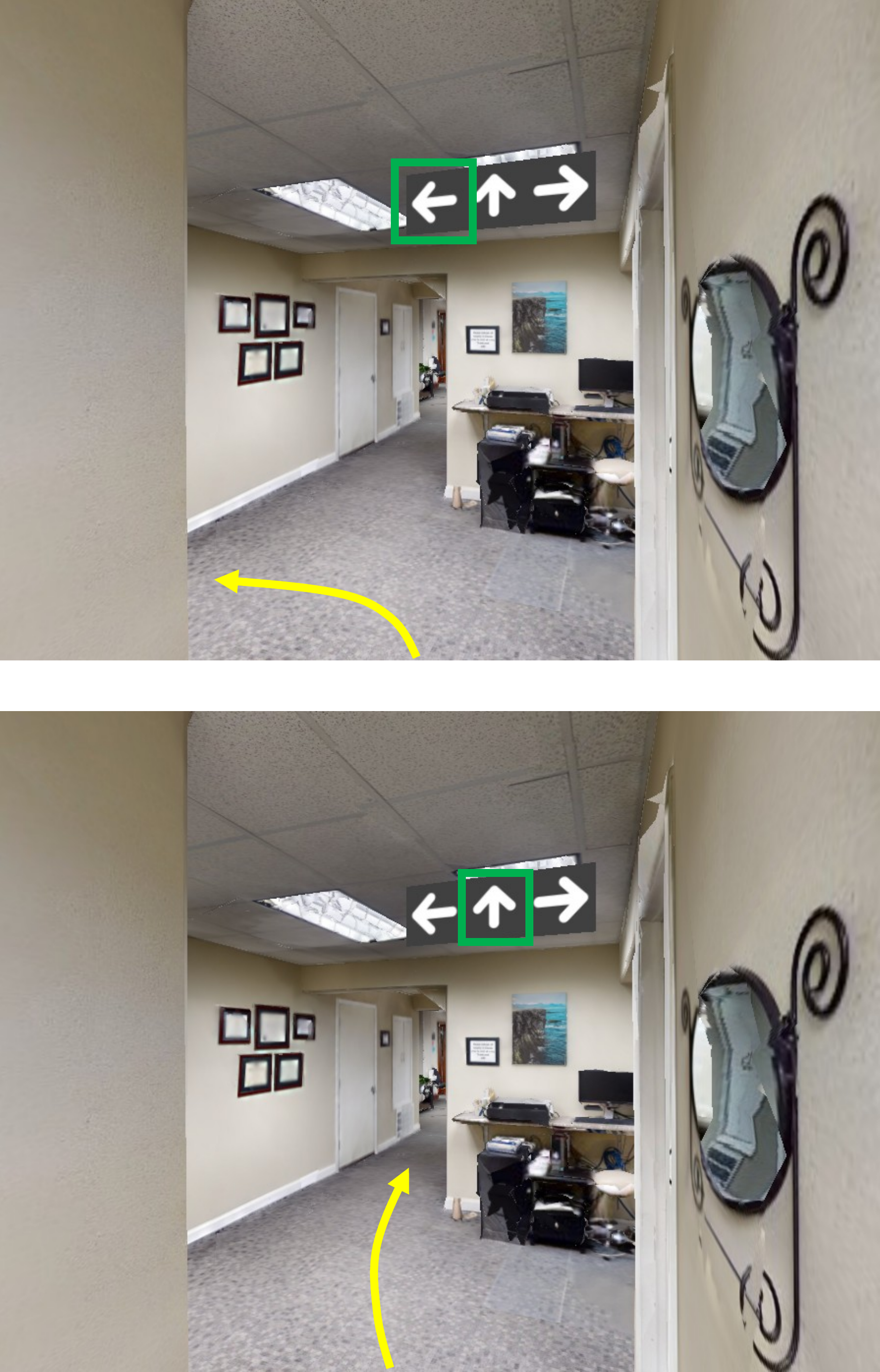}
        \end{minipage}
    }
    \subfigure[]
    {
        \begin{minipage}[b]{0.26\linewidth}
            \centering
            \includegraphics[width=1\linewidth]{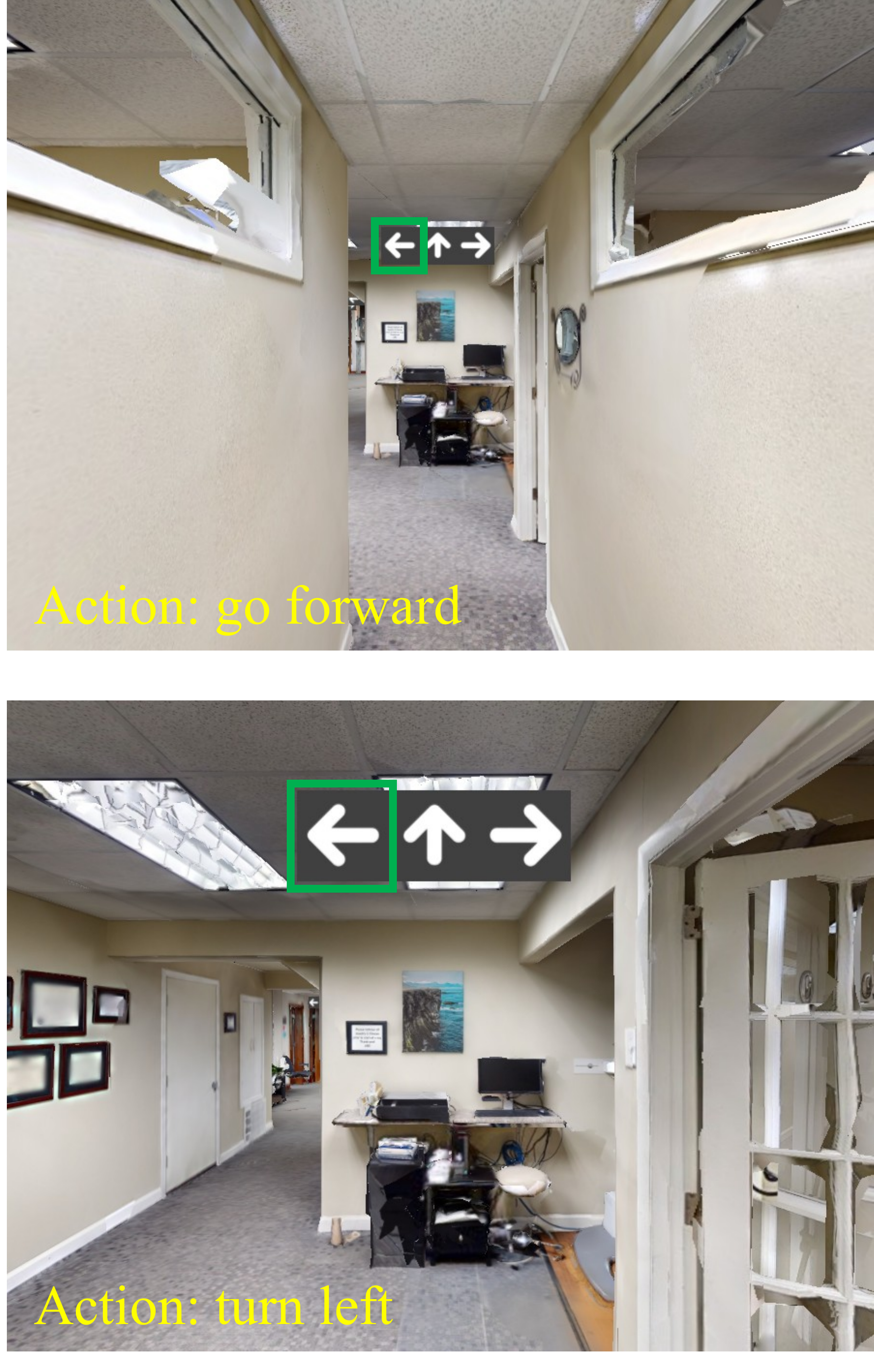}
        \end{minipage}
    }
    \subfigure[]
    {
        \begin{minipage}[b]{0.26\linewidth}
            \centering
            \includegraphics[width=1\linewidth]{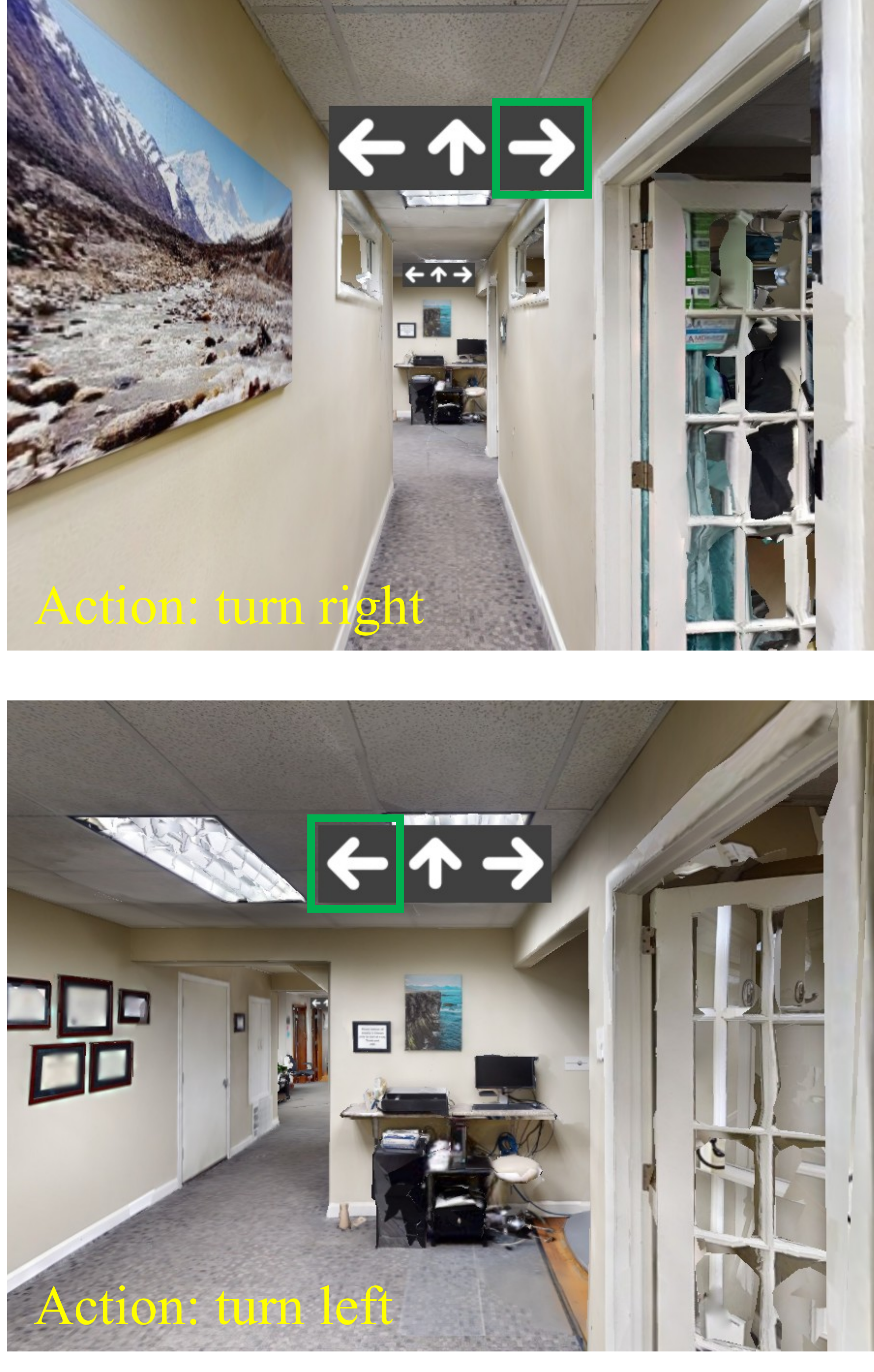}
        \end{minipage}
    }
    \caption{\textbf{Different action decisions under different circumstances.} (a) different semantic hints at the same location; (b) the same semantic hints at different locations; (c) different semantic hints at different locations.}
    \label{SignNav_7_8_9}
    \vspace{-8pt}
\end{figure}

Figure \ref{SignNav_7_8_9} illustrates the navigation action decisions made by the agent leveraging semantic hints of signage under different circumstances. It can be seen that different semantic hints and spatial locations of signage affect the agent's decisions. However, the semantic correspondence between the hint and the region of interest (RoI) in current observation remains invariant. To capture this correlation, we design a \textit{Spatial-Aware Transformer} module inspired by the Vision Transformer (ViT) \cite{dosovitskiy2020image}. The \textit{Spatial-Aware Transformer} takes as inputs the visual embedding $v^o_t$ and hint embedding $v^h_t$ at each step $t$, and then utilizes the self-attention mechanism to retrieve hint-correlated visual features from input embeddings.

Given the visual embedding $v^o_t$ and hint embedding $v^h_t$ from the previous stage, we first reshape the visual embedding into a sequence of flattened 2D patches $v^{patch}_t \in \mathbb{R}^{N\times (P^2\cdot C_o)}$, where $(H_o, W_o)$ is the resolution of the input visual embedding, $C_o$ is the number of channels, $(P, P)$ is the resolution of each embedding patch, and $N = H_oW_o/P^2$ is the resulting number of patches. These flattened patches are then projected to $D$ dimensions with trainable linear projection. In standard ViT, a learnable \textsf{[CLS]} token is initialized blindly to aggregate global information. In contrast, we explicitly initialize the \textsf{[CLS]} token using the projected hint embedding $v^h_t$. By treating the hint as the primary query token, the self-attention mechanism is forced to weigh visual patches based on their semantic relevance to the hint. The self-attention blocks in \textit{Spatial-Aware Transformer} is computed as follows:
\begin{equation}
    \boldsymbol{Q} = \boldsymbol{X}_{l-1}\boldsymbol{W}^Q_{l, k}, \boldsymbol{K} = \boldsymbol{X}_{l-1}\boldsymbol{W}^K_{l, k}, \boldsymbol{V} = \boldsymbol{X}_{l-1}\boldsymbol{W}^V_{l, k}
    \label{QKV}
\end{equation}
\begin{equation}
    \boldsymbol{H}_{l, k} = {\rm Softmax}\left(\frac{\boldsymbol{Q}\boldsymbol{K}^\top}{\sqrt{d_h}}\right)\boldsymbol{V}
    \label{softmaxQKV}
\end{equation}
where $\boldsymbol{W}^Q$, $\boldsymbol{W}^K$, and $\boldsymbol{W}^V \in \mathbb{R}^{hd_h \times d_h}$ are learnable linear projections specifically for queries, keys and values, $h$ is the number of attention heads and $d_h$ the attention head size of the network. The output of the \textit{Spatial-Aware Transformer} contains two distinct components enriched by cross-modal interaction. The output state corresponding to the patch sequence serves as the spatial embedding $f^{spa}_t$, where each patch feature is re-weighted by its responsiveness to the hint. Meanwhile, the output state of the hint token, which now aggregates the global context relevant to the navigation cue, is preserved to serve as the \textsf{[CLS]} token for the subsequent \textit{Temporal-Aware Transformer}.

\subsection{Temporal-Aware Transformer}
\begin{figure}[tp]
    \centering
    \includegraphics[width=0.8\linewidth]{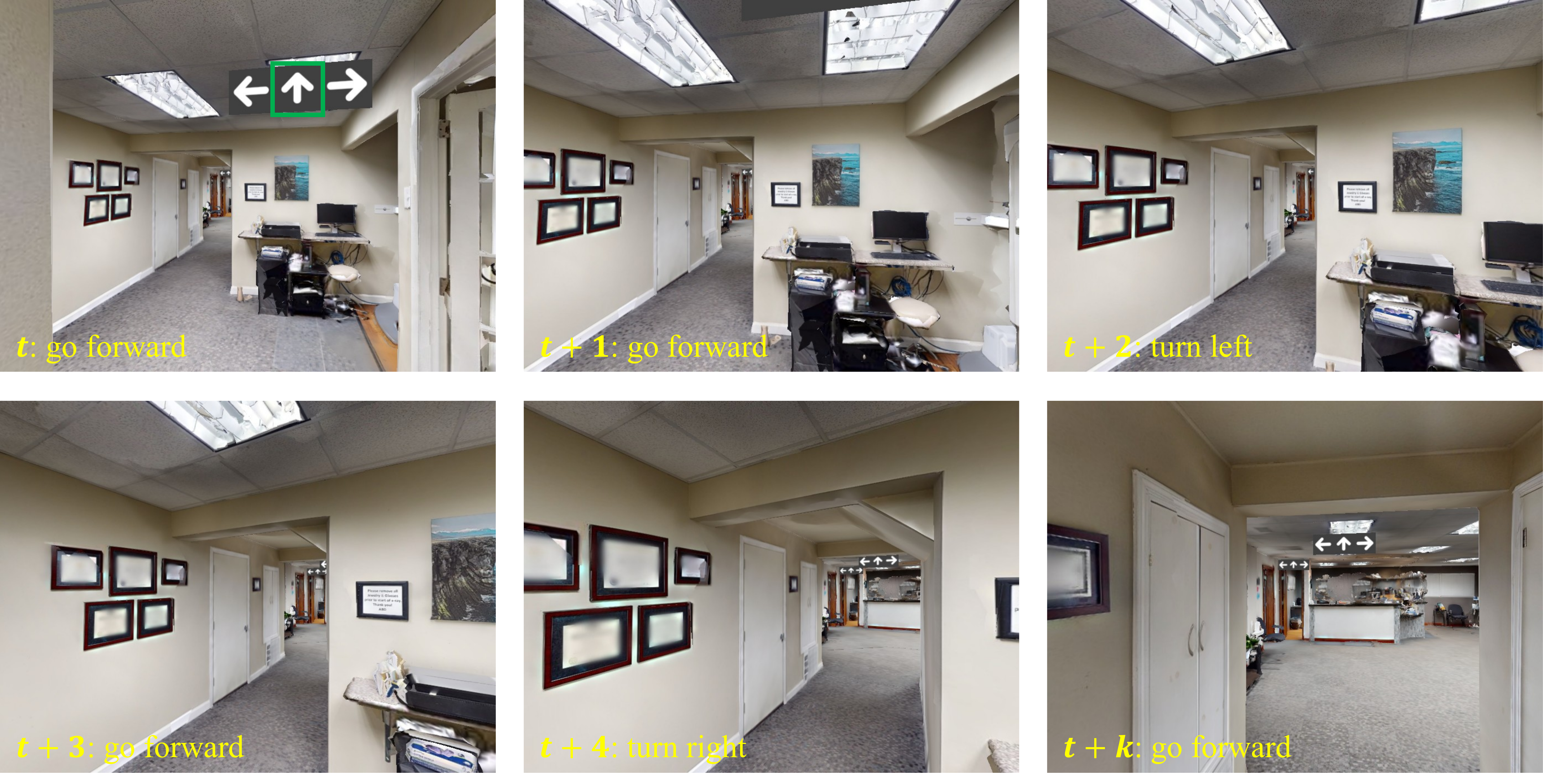}
    \caption{\textbf{One period in an episode that there is no visible semantic hints of signage.} The agent still needs to make correct decisions when there is no visible semantic hint.}
    \label{SignNav_10}
    \vspace{-8pt}
\end{figure}
In a long navigation episode of LSI environments, semantic hints of signage are not always visible. They often appear sparsely or become occluded due to limited field of view (as shown in Fig. \ref{SignNav_10}). When explicit guidance is absent, humans rely on episodic memory to recall the last observed hint and its spatial context to continue navigation. To address this, we propose a \textit{Temporal-Aware Transformer} that explicitly attends to a history of past states, enabling the agent to retrieve relevant semantic cues from previous steps directly. While signage is sparsely distributed throughout LSI environments, the semantic cues it provides evolve dynamically as the agent navigates, offering localized guidance. Consequently, rather than retaining an exhaustive navigational history, the agent only needs to maintain a fixed-length temporal window of recent states to make effective decisions.

Let $\mathcal{H}_k = \{ (f^{spa}_{t-k}, a_{t-k}), \dots, (f^{spa}_{t-1}, a_{t-1}) \}$ denote the history buffer of length $k$, containing the spatial embeddings from the \textit{Spatial-Aware Transformer} and the executed actions. To construct a compact state representation $s_i$ for each time step $i$, we first aggregate the spatial embedding $f^{spa}_i$ into a global vector $e^{spa}_i$ via average pooling. This visual vector is then fused with the action embedding $e^a_i$ (obtained via $a_i$ one-hot encoding followed by a linear projection) and normalized:
\begin{equation}
\begin{split}
    s_i = &{\rm LayerNorm}(e^{spa}_i\boldsymbol{W}^{spa}) + {\rm LayerNorm}(e^a_i\boldsymbol{W}^a)
\end{split}
\end{equation}
where $\boldsymbol{W}_s^{spa}$ and $\boldsymbol{W}_s^a$ are learnable projection matrices.

The sequence of state embeddings ${s_{t-k}, \dots, s_{t-1}}$ serves as the input to the \textit{Temporal-Aware Transformer}. This module comprises $L$ layers of Transformer blocks. Each block consists of a Multi-Head Self-Attention (MSA) layer followed by a Feed-Forward Network (FFN), with residual connections and layer normalization applied at each sub-layer. The $k$-th attention head at the $l$-th layer performs self-attention over $\boldsymbol{X}_{l-1}$ following \eqref{QKV} and \eqref{softmaxQKV}. The outputs from all the attention heads will be concatenated and projected onto the same dimension as the input as:
\begin{equation}
    \boldsymbol{H}_l = \left[\boldsymbol{H}_{l, 1};\ldots ;\boldsymbol{H}_{l, k}\right]\boldsymbol{W}^o_l
\end{equation}
where $k$ is the total number of attention heads, $[;]$ denotes concatenation and $\boldsymbol{W}^O\in \mathbb{R}^{hd_h\times hd_h}$ is a learnable linear projection. Finally, the output of layer $l$ is formulated by:
\begin{align}
    \boldsymbol{H}^\prime_l &= {\rm LayerNorm}(\boldsymbol{H}_l + \boldsymbol{X}_{l-1}) \\
    \boldsymbol{X}^\prime_l &= {\rm GeLU}(\boldsymbol{H}^\prime_l\boldsymbol{W}^{F_1}_l)\boldsymbol{W}^{F_2}_{l} \\
    \boldsymbol{X}_l &= {\rm LayerNorm}(\boldsymbol{H}^\prime_l + \boldsymbol{X}^\prime_{l})
\end{align}
where ${\rm GeLU}$ is the Gaussian Error Linear Unit activation function \cite{hendrycks2016gaussian}. The final output vector $f_{t}^{s}$ (corresponding to the last token in the sequence) aggregates the historical context and the hint-grounded spatial features. This vector encapsulates the agent's current belief state and is passed to the policy network to predict the next navigable action.

\subsection{Training Objective}
We formulate the navigation task as a sequential decision-making problem. Based on the comprehensive state representation $f_{t}^{s}$, the policy is parameterized by a fully connected network (FCN) to predict the action distribution $p^a_t$ over the discrete action space:
\begin{equation}
    p^a_t = {\rm Softmax}(f_{t}^{s} \boldsymbol{W}^{s})
\end{equation}
where $\boldsymbol{W}^{s}$ denotes the learnable parameter matrix. The primary objective is to maximize the likelihood of the ground-truth trajectories. To achieve this, we employ teacher-forcing imitation learning with inflection weighting \cite{wijmans2019embodied}, which is helpful for navigation with long sequences of repeated actions (e.g., traversing a long hallway). Formally, we minimize the weighted cross-entropy loss function, defined as:
\begin{equation}
    \mathcal{L}_{loss} = -\sum_t w_t a^\ast_t \log(p^a_t)
\end{equation}
where $p^a_t$ represents the predicted action distribution, $a^\ast_t$ is the ground-truth teacher action, and $w_t$ is the inflection weight coefficient at step $t$.

Standard imitation learning in auto-regressive settings typically suffers from covariate shift, where agents are not exposed to the consequences of their own errors during training. To address this issue, we apply Dataset Aggregation (DAgger) \cite{ross2011reduction} to train on the aggregated set of trajectories from all iterations $1$ to $n$. As demonstrated in the ablation study, this strategy significantly enhances the generalization capability of the agent.
\section{Experiments}
\subsection{Experimental setup}
\textbf{Dataset.} We train and evaluate our approach using the Habitat simulator on our LSI-Dataset. The scene dataset is split into train, val-seen and val-unseen sets with 16, 16 and 4 environments, respectively. Environments in val-seen split are the same as training, while environments in val-unseen split are different from training. We collect 8000, 3200 and 800 episodes data from each split, and the average trajectory length is about 10 $m$.

\textbf{Baselines.} We evaluate our START model against three distinct paradigms that leverage signage for navigation: rule-based method \cite{liang2020no}, VLM-based approach, and ViNT \cite{Shah2023ViNTAF}. Specifically, the \textbf{Rule-based} method \cite{liang2020no} relies on hand-crafted rules to map semantic hints to topological map nodes. For the \textbf{VLM-based} baseline, since no existing work directly applies VLMs to end-to-end signage-based navigation, we develop a custom agent using Qwen3 \cite{qwen3}. This open-source model provides robust reasoning capabilities while circumventing the prohibitive costs of querying closed-source APIs for sequential decision-making.The prompt context is available if requested. Finally, \textbf{ViNT} \cite{Shah2023ViNTAF}, a visual navigation foundation model, is adapted to the SignNav task by replacing its sub-goal encoder with a hint image encoder to output four discrete actions. For a fair comparison, we retrain ViNT on our LSI-Dataset, employing early stopping based on validation performance.

\textbf{Metrics.} For all approaches, we adopt the following standard evaluation metrics described by \cite{anderson2018evaluation} and \cite{Ilharco2019GeneralEF}: 1) \textit{Success Rate} (SR); 2) \textit{normalized Dynamic Time Warping} (NDTW); 3) \textit{Success weighted by normalized Dynamic Time Warping} (SDTW). SR means the ratio of episodes reaching the destination with a maximum error of 1 $m$ to the target location. NDTW measures similarity between the entirety of agent trajectory and ground-truth trajectory, softly penalizing deviations. It naturally reflects the performance of algorithms by forcing the alignment between ground-truth and robot trajectories. SDTW (SR$\times$NDTW) stresses the importance of reaching the goal, where SDTW equals to NDTW if the episode is successful and zero otherwise.
\begin{figure}[tp]
    \centering
    \includegraphics[width=0.9\linewidth]{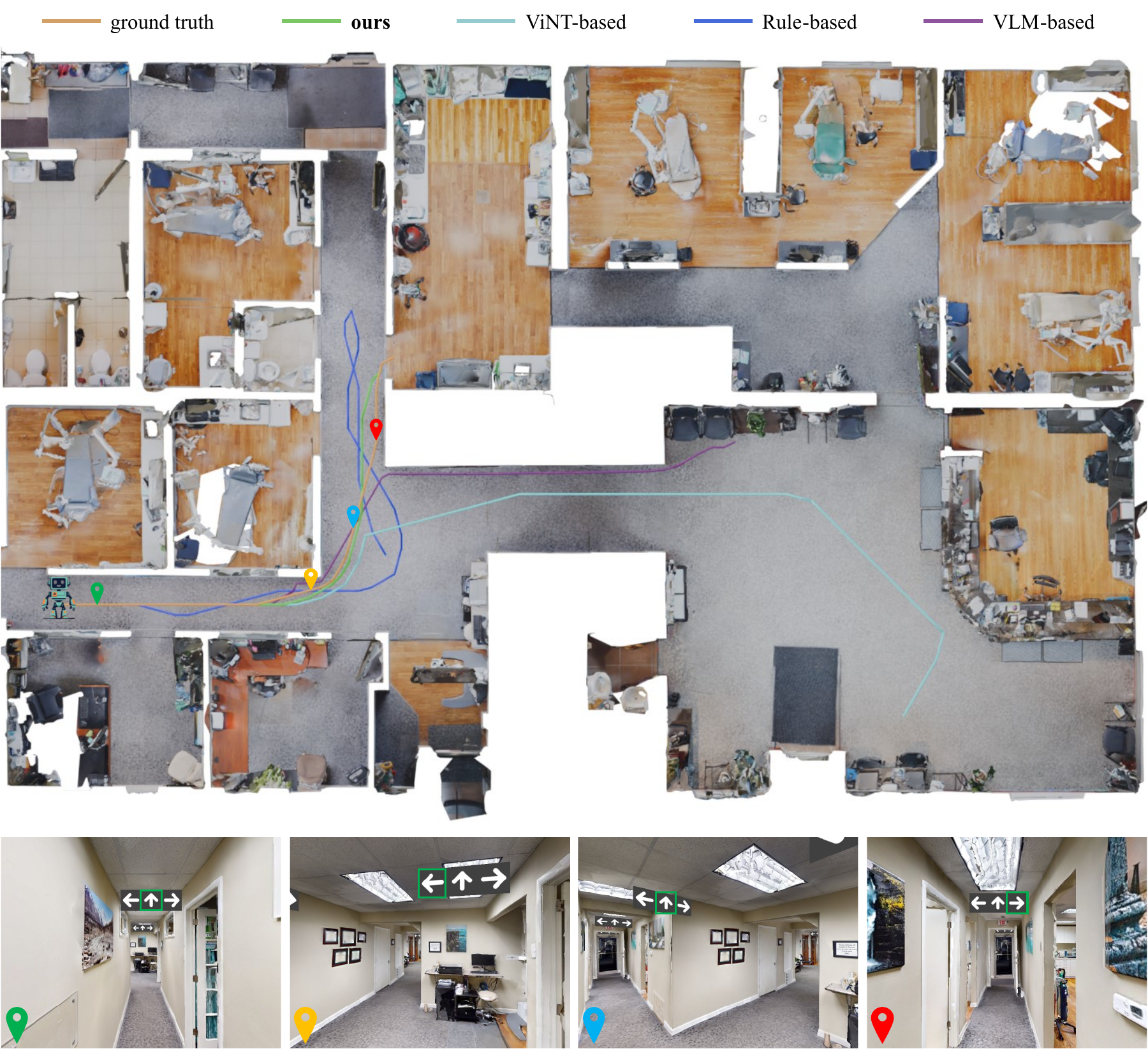}
    \caption{\textbf{Trajectories visualization of our method compared to baselines.} The second row shows the first-person-view of agent as it navigates along the ground-truth trajectory.}
    \label{SignNav_20}
    \vspace{-8pt}
\end{figure}

\textbf{Implementation details.} For our START model, we first use ResNet-50 to encode the hint image of current observation to get the semantic hint embedding. We adopt the ViT-B/32 \cite{dosovitskiy2020image} as the backbone of \textit{Spatial-Aware Transformer} and replace the \textsf{[CLS]} token in raw ViT-B/32 architecture with semantic hint embedding. For the \textit{Temporal-Aware Transformer}, we use the BERT-like architecture \cite{devlin2019bert}. The hidden embedding size, depth of attention layers and number of attention heads both in spatial-aware and temporal-aware Transformers are set to 768, 6 and 6, respectively. A simple MLP prediction head is used to predict the final action distribution. In the training process, we first train our START model with teacher-forcing and then fine-tune it using the DAgger algorithm for 10 iterations. Each iteration is trained for 100 epochs with fixed learning rate of 1.0e-5 and batch size of 12 on 6 NVIDIA RTX4090 GPUs. All the evaluation experiments are completed using 1 NVIDIA RTX4090 GPU.

\subsection{Main Results}
\begin{table}[tp]
    \centering
    \caption{Performance comparison with different baselines.}
    \label{tab:comparison}
    \scalebox{0.8}
    {
    \begin{tabular}{ccccccc}
        \toprule
        \multirow{2}{*}{Methods} & \multicolumn{3}{c}{Val-Seen} & \multicolumn{3}{c}{Val-Unseen} \\
        \cmidrule(lr){2-4} \cmidrule(lr){5-7}
         & SR & NDTW & SDTW & SR & NDTW & SDTW \\
        \midrule
        \multicolumn{1}{c}{VLM-based} & 0.28 & 0.46 & 0.21 & 0.30 & 0.54 & 0.25 \\
        \multicolumn{1}{c}{Rule-based \cite{liang2020no}} & 0.51 & 0.65 & 0.42 & 0.67 & 0.73 & 0.56 \\
        \multicolumn{1}{c}{ViNT \cite{Shah2023ViNTAF}} & 0.76 & 0.71 & 0.64 & 0.68 & 0.69 & 0.57 \\
        \midrule
        \multicolumn{1}{c}{\textbf{START}} & \textbf{0.90} & \textbf{0.80} & \textbf{0.76} & \textbf{0.80} & \textbf{0.74} & \textbf{0.67} \\
        \bottomrule
    \end{tabular}
    }
    \vspace{-8pt}
\end{table}

\textbf{How well can START perform the SignNav task in LSI environments?} Table \ref{tab:comparison} summarizes the evaluation on the LSI-Dataset. START significantly outperforms all baselines across all metrics, achieving absolute gains of +12\% in SR and +5\% in NDTW over the strongest baseline, ViNT, on the val-unseen split. Although retrained on our dataset, ViNT relies on basic CNN encoders that lack explicit spatial grounding of semantic hints and struggle to capture temporal state dynamics. The Rule-based method, the only approach requiring online odometry for real-time mapping and localization, unexpectedly performs better on the val-unseen split (+16\% SR, +8\% NDTW), which we attribute to data distribution biases within the statistically larger val-seen split. Conversely, the VLM-based method yields the poorest performance (30\% SR, 0.54 NDTW on val-unseen). While multi-modal VLMs excel at abstract visual reasoning, they still struggle with grounding physical interactions, particularly when directly predicting low-level control commands. Qualitative trajectory comparisons are provided in Fig. \ref{SignNav_20}.
\begin{table}[bp]
    \centering
    \caption{Trajectory error statistics at different distances.}
    \label{tab:trajectory error}
    \scalebox{0.8}
    {
    \begin{tabular}{cccccc}
        \toprule
        \multirow{2}{*}{\textbf{RMSE} $(m)$} & \multicolumn{5}{c}{Navigation Distance $(m)$} \\
        \cmidrule{2-6}
         & 40 & 80 & 120 & 160 & 200 \\
        \midrule
        \multicolumn{1}{c}{Val-Seen} & 0.28 & 0.25 & 0.24 & 0.25 & 0.24 \\
        \multicolumn{1}{c}{Val-Unseen} & 0.25 & 0.26 & 0.26 & 0.28 & 0.27 \\
        \bottomrule
    \end{tabular}
    }
\end{table}
\begin{figure*}[tp]
    \centering
    \includegraphics[width=0.9\linewidth]{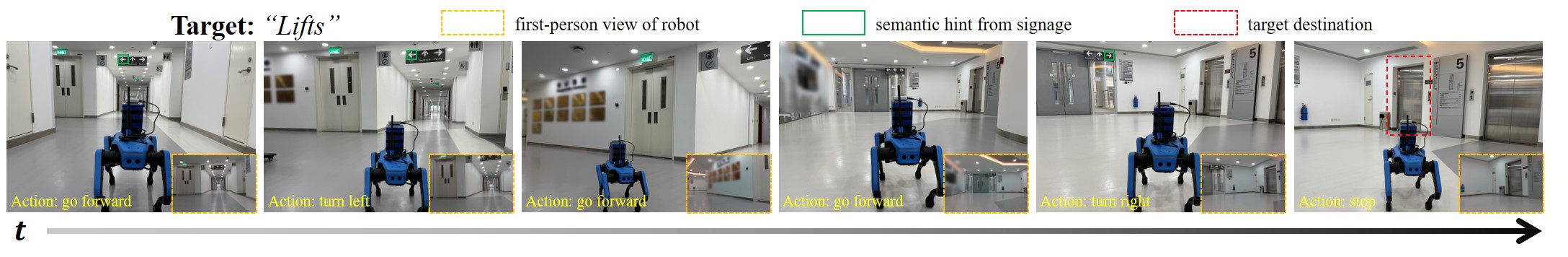}
    \caption{Real-world deployment of our method in an office building to find the \textit{``Lifts"} leveraging signage.}
    \label{SignNav_14}
\end{figure*}

\textbf{How long can our START model navigate in the SignNav task?} While existing embodied navigation tasks primarily focus on household environments with short episodes (e.g., \(\sim\)10 \(m\)), SignNav targets LSI spaces like hospitals, where real-world trajectories can span hundreds of meters. To evaluate our START model's long-horizon robustness, we tested it on representative scenes from both val-seen and val-unseen splits, using episodes ranging from 40 to 200 \(m\). START successfully completed all episodes. As shown in Table \ref{tab:trajectory error}, the absolute trajectory error (\textbf{RMSE}) remains remarkably stable, approximately 0.25 \(m\) (val-seen) and 0.27 \(m\) (val-unseen), regardless of the navigation distance. This length-agnostic performance stems from the fact that, unlike VLN which requires continuous agent-instruction localization, SignNav relies on localized semantic cues. This property allows our \textit{Temporal-Aware Transformer} to function effectively over extended periods without accumulating localization errors.
\begin{table}[tp]
    \centering
    \caption{Ablation study for spatial and temporal-aware modules.}
    \label{ablation study for spatial-temporal}
    \scalebox{0.8}
    {
    \begin{tabularx}{\linewidth}{XXcccccc}
        \toprule
        \multirow{2}{*}{\makecell{Spa- \\ Aware}} & \multirow{2}{*}{\makecell{Tem- \\ Aware}} & \multicolumn{3}{c}{Val-Seen} & \multicolumn{3}{c}{Val-Unseen} \\
        \cmidrule(lr){3-5} \cmidrule(lr){6-8} 
         &  & SR & NDTW & SDTW & SR & NDTW & SDTW \\
        \midrule
        \multicolumn{1}{c}{$\times$} & \multicolumn{1}{c}{$\checkmark$} & 0.60 & 0.67 & 0.50 & 0.39 & 0.57 & 0.33 \\
        \multicolumn{1}{c}{$\checkmark$} & \multicolumn{1}{c}{$\times$} & 0.61 & 0.66 & 0.51 & 0.47 & 0.59 & 0.39 \\
        \multicolumn{1}{c}{$\checkmark$} & \multicolumn{1}{c}{$\checkmark$} & \textbf{0.82} & \textbf{0.76} & \textbf{0.69} & \textbf{0.72} & \textbf{0.72} & \textbf{0.60} \\
        \bottomrule
    \end{tabularx}
    }
\end{table}
\begin{figure}[tp]
    \centering
    \subfigure[]
    {
        \begin{minipage}[b]{0.25\linewidth}
            \centering
            \includegraphics[width=1\linewidth]{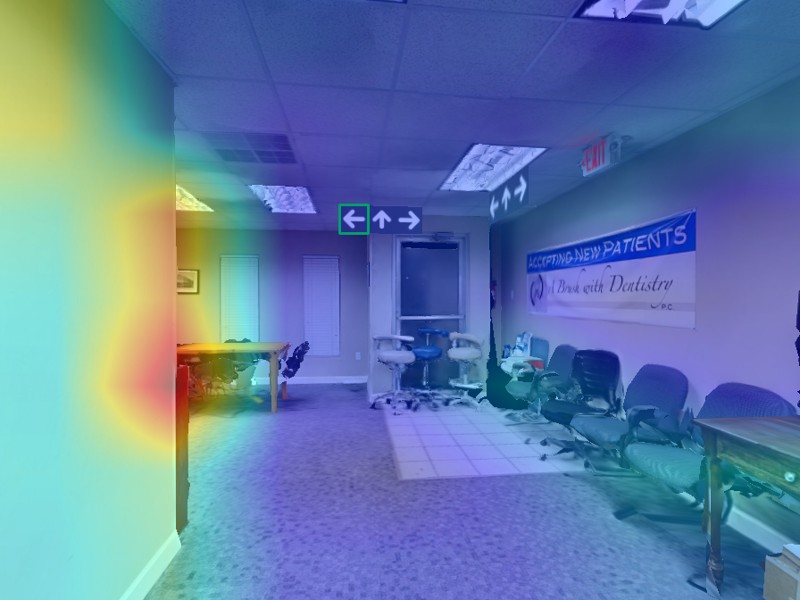}
        \end{minipage}
    }
    \subfigure[]
    {
        \begin{minipage}[b]{0.25\linewidth}
            \centering
            \includegraphics[width=1\linewidth]{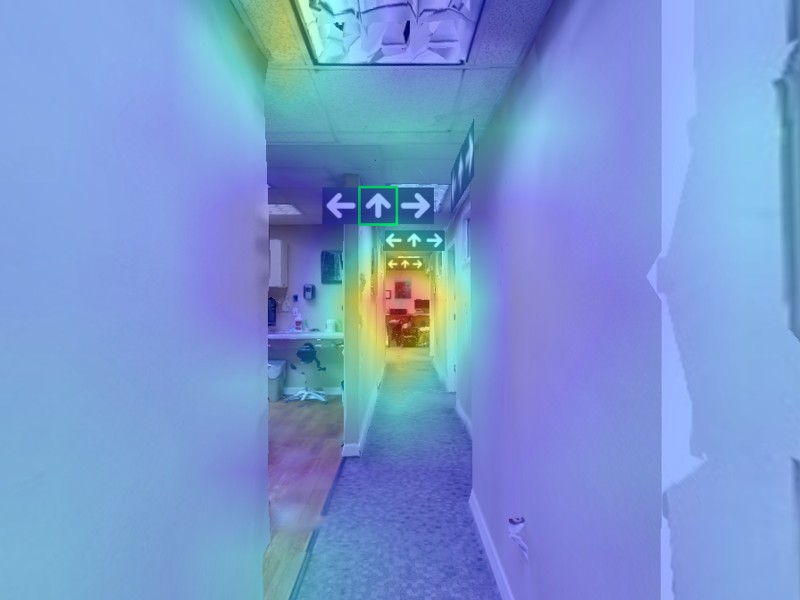}
        \end{minipage}
    }
    \subfigure[]
    {
        \begin{minipage}[b]{0.25\linewidth}
            \centering
            \includegraphics[width=1\linewidth]{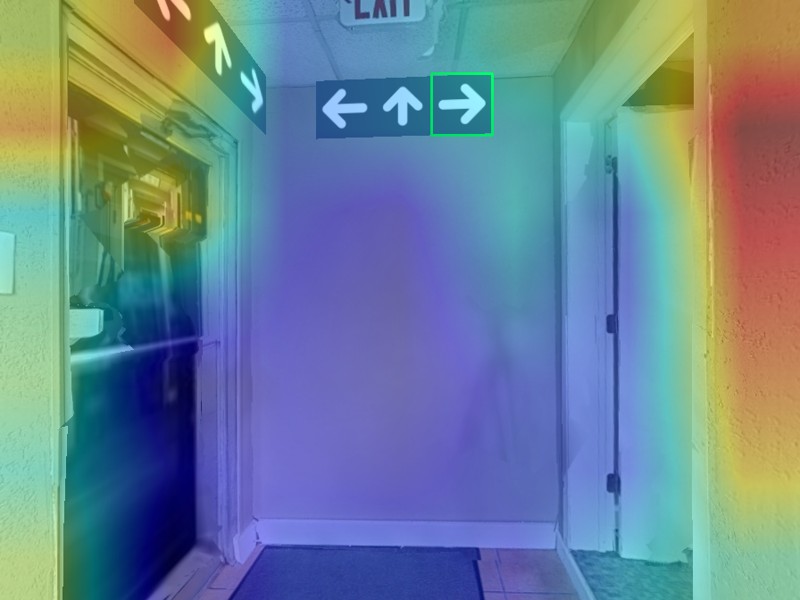}
        \end{minipage}
    }
    \caption{\textbf{Attention map visualization both with the spatial-aware and temporal-aware modules in our START model.} The darker the color, the greater the attention weight.}
    \label{SignNav_15_16_17}
    \vspace{-8pt}
\end{figure}

\textbf{Can our START model be deployed successfully in the real world?} To validate START's practical applicability for signage-based navigation, we deploy it on a DeepRobotics Lite2 robot (Fig. \ref{SignNav_14}). The robot captures RGB-D observations via an Intel RealSense D435i camera, streaming them over a wireless LAN to a remote desktop equipped with one NVIDIA RTX4090 GPU. To mitigate the camera's noisy and limited depth sensing, we employ DepthAnything-V2 \cite{yang2024depth} to generate high-fidelity depth images. Concurrently, a locally deployed Qwen3-8B model handles the perception and parsing of semantic hints, and also determines episode termination at each step, offering an optimal balance of inference speed and accuracy. By integrating the RGB image, generated depth image, hint image and its position, START predicts discrete navigation actions that are transmitted back to the robot at a frequency of \SI{0.5}{\hertz}. Real-world deployment videos are available in the multimedia material.

\subsection{Ablation Studies}
\textbf{How important is the spatial-temporal aware module for SignNav task?} To assess the individual contributions of the spatial- and temporal-aware modules in addressing the core challenges of SignNav, we train variants of our model from scratch without DAgger (Table \ref{ablation study for spatial-temporal}). While both the temporal-only (Row 1) and spatial-only (Row 2) baselines achieve comparable performance on the val-seen split, the spatial-only model outperforms its temporal counterpart on val-unseen (+8\% SR, +2\% NDTW). Because the spatial-only model makes single-step predictions independent of historical states, it relaxes input distribution constraints, thereby improving generalization in unseen environments. Ultimately, integrating both modules (Row 3) yields the best overall performance, delivering substantial relative gains of +25\% SR and +13\% NDTW on the val-unseen split. Furthermore, visualizations of the final attention layer (Fig. \ref{SignNav_15_16_17}) corroborate that the synergy of both modules is critical for effective navigation.

\textbf{How much does training with DAgger help?} In the training process of our START model, we apply two-stage training strategy. The first stage uses teacher-forcing to train a prior policy and then the DAgger algorithm is used to finetune it. Table \ref{ablation for DAgger} presents the evaluation results of our model trained w/o and w/ DAgger. Different from simple data augmentation techniques, training with DAgger enables agent to collect more interactive data with environments, which results in better performance by relative 8\% and 8\% improvements in SR on val-seen and val-unseen splits, respectively.
\begin{table}[tp]
    \centering
    \caption{Ablation study for the impact of training with DAgger.}
    \label{ablation for DAgger}
    \scalebox{0.8}
    {
    \begin{tabular}{ccccccc}
        \toprule
        \multirow{2}{*}{DAgger} & \multicolumn{3}{c}{Val-Seen} & \multicolumn{3}{c}{Val-Unseen} \\
        \cmidrule(lr){2-4} \cmidrule(lr){5-7} 
         & SR & NDTW & SDTW & SR & NDTW & SDTW \\
        \midrule
        $\times$ & 0.82 & 0.76 & 0.69 & 0.72 & 0.72 & 0.60 \\
        $\checkmark$ & \textbf{0.90} & \textbf{0.80} & \textbf{0.76} & \textbf{0.80} & \textbf{0.74} & \textbf{0.67} \\
        \bottomrule
    \end{tabular}
    }
    \vspace{-8pt}
\end{table}
\indent\textbf{What is the impact of using different input modalities?} Table \ref{ablation for modal inputs} evaluates the impact of different visual inputs on START's performance. The RGB-only variant yields the poorest results, often causing the agent to exhibit oscillatory behavior (swaying) during navigation. Conversely, relying solely on depth observations for navigation (while retaining RGB for hint images) achieves the best overall performance. Surprisingly, this depth-only configuration outperforms the combined RGB-D model, yielding relative SR improvements of +4\% and +5\% on the val-seen and val-unseen splits, respectively. We attribute this to the inherent robustness of depth data against illumination and texture variations, which facilitates better generalization across environments. Nevertheless, RGB inputs remain indispensable in our final pipeline for extracting semantic hints from signage and determining episode termination.

\section{Conclusion}
This paper introduces SignNav, a novel embodied navigation task where agents navigate LSI environments leveraging semantic hints from signage. To facilitate this research, we formalize the task and construct the LSI-Dataset across 20 diverse environments. To tackle the inherent challenges of SignNav, we propose the START model, which explicitly grounds semantic hints within current visual observation and captures temporal dynamics across historical states. Extensive evaluations on the LSI-Dataset demonstrate that START achieves state-of-the-art performance, significantly outperforming strong baselines. Finally, successful real-world deployments on a Lite2 robot validate the practical efficacy of our approach.
\begin{table}[tp]
    \centering
    \caption{Ablation study for the impact of different modal inputs.}
    \label{ablation for modal inputs}
    \scalebox{0.8}
    {
    \begin{tabularx}{\linewidth}{XXcccccc}
        \toprule
        \multirow{2}{*}{RGB} & \multirow{2}{*}{Depth} & \multicolumn{3}{c}{Val-Seen} & \multicolumn{3}{c}{Val-Unseen} \\
        \cmidrule(lr){3-5} \cmidrule(lr){6-8} 
         &  & SR & NDTW & SDTW & SR & NDTW & SDTW \\
        \midrule
        \multicolumn{1}{c}{$\checkmark$} & \multicolumn{1}{c}{$\times$} & 0.67 & 0.70 & 0.56 & 0.59 & 0.68 & 0.49 \\
        \multicolumn{1}{c}{$\times$} & \multicolumn{1}{c}{$\checkmark$} & \textbf{0.86} & \textbf{0.78} & \textbf{0.73} & \textbf{0.77} & \textbf{0.73} & \textbf{0.64} \\
        \multicolumn{1}{c}{$\checkmark$} & \multicolumn{1}{c}{$\checkmark$} & 0.82 & 0.76 & 0.69 & 0.72 & 0.72 & 0.60 \\
        \bottomrule
    \end{tabularx}
    }
    \vspace{-8pt}
\end{table}

\bibliographystyle{IEEEtranBST/IEEEtran}
\bibliography{root}
\addtolength{\textheight}{-12cm}   

\end{document}